\documentclass[%
 reprint,
 amsmath,amssymb,
 aps,
]{revtex4-2}

\usepackage{graphicx}
\usepackage{dcolumn}
\usepackage{bm}

\usepackage{xcolor}
\usepackage{soul}

\newcommand{\XX}{\mathbf{X}}

\newcommand{\FF}{\mathbf{F}}

\begin{document}

\preprint{APS/123-QED}

\title{Physics-Based Deep Neural Networks for \\Beam Dynamics in Charged Particle Accelerators}

\author{Andrei Ivanov}
\altaffiliation{05x.andrey@gmail.com, andrei.ivanov@desy.de\\Phys. Rev. Accel. Beams 23, 074601 – Published 7 July 2020\\
DOI: 10.1103/PhysRevAccelBeams.23.074601\\
https://link.aps.org/doi/10.1103/PhysRevAccelBeams.23.074601
}
\author{Ilya Agapov}
\affiliation{Deutsches Elektronen Synchrotron DESY, Notkestrasse 85, 22607, Hamburg, Germany.}






\date{\today}

\begin{abstract}
This paper presents a novel approach for constructing neural networks which model charged particle beam dynamics. In our approach, the Taylor maps arising in the representation of dynamics are mapped onto the weights of a polynomial neural network. The resulting network approximates the dynamical system with perfect accuracy prior to training and provides a possibility to tune the network weights on additional experimental data. We propose a symplectic regularization approach for such polynomial neural networks that always restricts the trained model to Hamiltonian systems and significantly improves the training procedure. The proposed networks can be used for beam dynamics simulations or for fine-tuning of beam optics models with experimental data. The structure of the network allows for the modeling of large accelerators with a large number of magnets. We demonstrate our approach on the examples of the existing PETRA III and the planned PETRA IV storage rings at DESY. 
\end{abstract}

\maketitle


\section{\label{sec:level1}Introduction}

Machine Learning (ML) techniques are finding increasing usage in various aspects of particle accelerators, including fault prediction, performance optimization, and virtual diagnostics. For more information see e.g. \cite{edelen2018opportunities}. Many applications could benefit from having a beam optics model which on the one hand accurately represents the beam dynamics of the accelerator, and on the other hand, can be trained or adjusted on the limited amount of experimental data to serve as a model of the real machine with various imperfections. Machine Learning methods and neural networks (NN) in particular hold a promise for constructing such models.

However, applying ML methods for learning the behavior of dynamical systems such as those describing linear and nonlinear dynamics in charged particle accelerators can in some cases require a prohibitively large amount of training data. There is a need for ML-based modeling approaches that reduce reliance on large data sets. 
A major concern is that models trained with limited observations may not generalize well beyond the training dataset and will thus exhibit poor predictive power. A significant improvement potential in the generalization ability and the training performance of ML models such as NNs lies in incorporating physics constraints either in the NN architecture or in the training process.

NNs are often employed for  system learning and control \cite{Nagabandi,ChenY}, when models are trained either with large measured or simulated datasets. By physics-inspired neural networks \cite{Thuerey} authors generally mean either incorporating domain knowledge in the traditional NN or providing additional loss functions to ensure physically consistent predictions \cite{PINN, HNN1, HNN2}. Most of the authors either incorporate numerical integrators for ordinary differential equations (ODEs) into NN or, oppositely, parametrize equations with NNs.
There also exist various studies related to PNNs in the literature \cite{ref10,ref11,ref12}. So, in \cite{ref10,ref11} polynomial architectures that approximate differential equations are discussed. All studies related to PNNs regard the polynomial architectures as black-box models, and they do not indicate the architectures' connection to the ODEs.

We propose to incorporate the physics constraints on the single-particle beam dynamics in accelerators by introducing NN architecture which is directly derived from the Taylor maps corresponding to the accelerator components and additionally imposing symplectic constraints on the network. 
The approach presented in the paper is based on works \cite{MLM, TMPNN, LiePNN}, where the authors target building polynomial neural networks (PNN) that approximate the exact system of ODEs. In contrast to these works, we do not target solving exact equations and simplified physical problems, but focus on the practical application of the fine-tuning of the PNN with measured data.

In this paper, we first introduce the polynomial neural network (PNN) architecture incorporating Taylor map (TM-PNN) information. Training and symplectic regularization are then discussed, followed by the application of TM-PNNs to simulating beam dynamics in storage rings. Finally, the results of fine-tuning the TM-PNN based on experimental data at the PETRA III \cite{Balewski:2004iz} storage ring and simulations for the proposed PETRA IV \cite{Schroer:ig5056} ring at DESY are presented.

\begin{figure}[ht]
\includegraphics[width=0.4\textwidth]{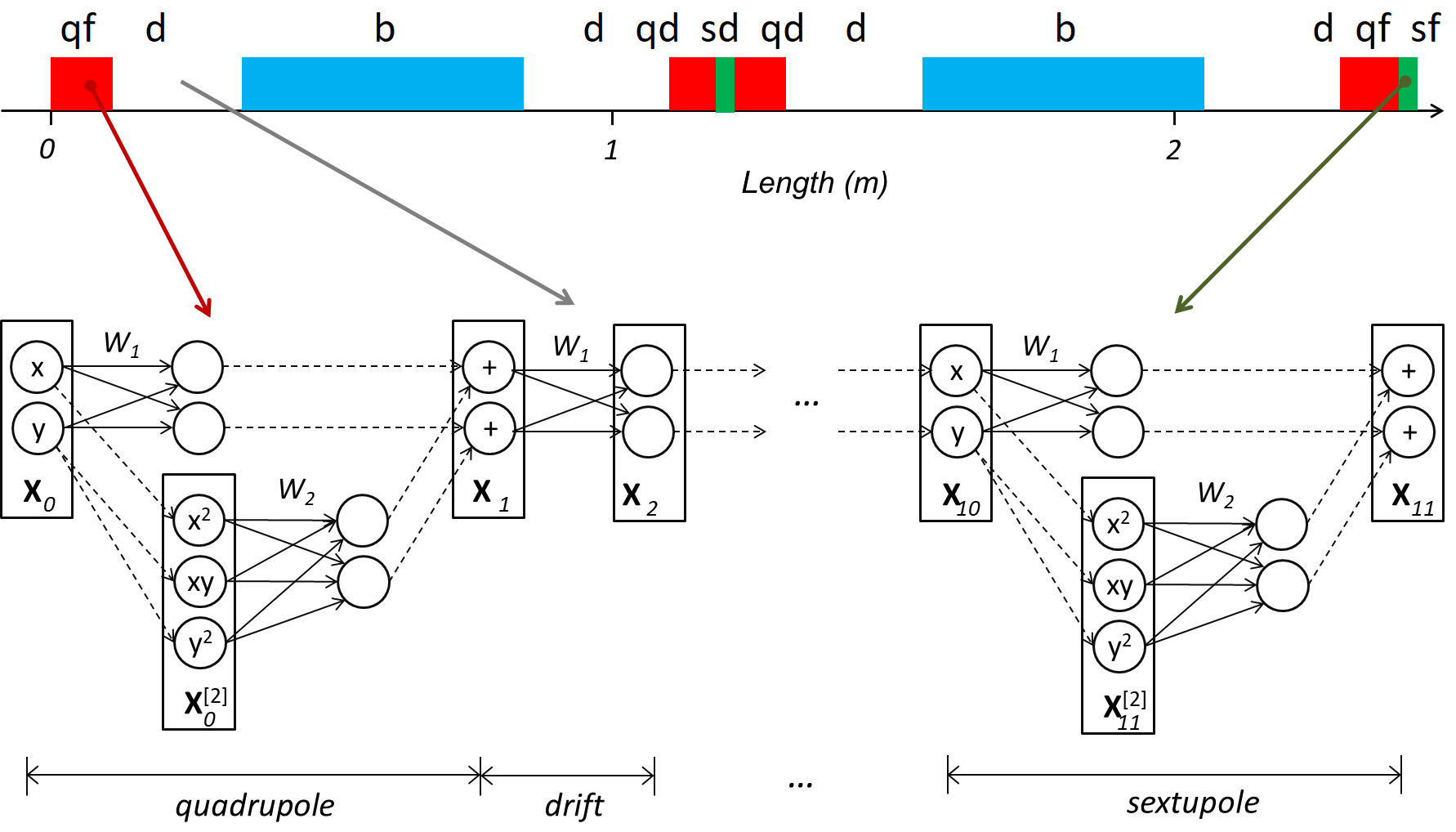}
\caption{Schematic representation of a FODO structure with eight magnets and a corresponding TM-PNN. The lattice consists of drifts (d), bending magnets (b), shown in blue, focusing (qf) and defocusing (qd) quadrupoles, shown in red, and focusing (sf) and defocusing (sd) sextupoles, shown in green.}
\label{fig.02}
\end{figure}

\section{Polynomial Neural network incorporating Taylor maps}

Taylor maps are a natural and well-known approach of treating beam dynamics in particle accelerators, see e.g. \cite{SLAC-PUB-9574, ref13}. A Taylor map $\mathcal M $ is a polynomial transformation of the phase space coordinates of a particle in the form
\begin{equation}
	\label{tmap}
	\XX(s) \equiv \XX_1 = W_0 + W_1\,\XX_0+W_2\,\XX_0^{[2]}+\ldots+W_k\,\XX_0^{[k]}.
\end{equation}
Here $\XX_{0,1} \in R^n$ are the n-dimensional phase space coordinates of a particle, $W_i$ are weight matrices, and $\XX^{[k]}$ are the $k$-th Kronecker powers of vector $\XX$. For example, for $\XX = (x_1, x_2)$,  $\XX^{[2]} = (x_1^2, x_1x_2, x_2^2)$, $\XX^{[3]} = (x_1^3, x_1^2x_2, x_1x_2^2$, $x_2^3)$ and so on. The weights of \eqref{tmap} can be derived from equations of particle motion that can be described in a general form with polynomial coefficients
\begin{equation}
\label{odesystem}
\frac{d}{ds}\XX = \FF(s, \XX) = \sum_{j=0}^{n} P_{k}(s)\XX^{[k]},
\end{equation}
where $s$ is an independent variable, for which in accelerator physics the path length of the reference particle is usually taken. Differentiating \eqref{tmap} and combining it with \eqref{odesystem}  yields an equation for weight matrices $W_i$
\begin{equation}
\frac{d}{ds}W_i = f_i(W_0, \ldots, W_k, P_0, \ldots P_n),\; i=\overline{0,k},
\label{eq:dW}
\end{equation}
where $f_i$ are functions of matrices $W_i$ and $P_i$. For instance, $f_0 = P_0 + P_1W_0 + \ldots P_k W_0^{[k]}$. Solving \eqref{eq:dW} for $W_i$ gives the dynamics of the system for all initial conditions by means of \eqref{tmap}. The success of this technique for modeling beam dynamics in accelerators lies in the fact that Taylor maps for individual magnets can be readily constructed to desired order, and long sequences of magnets comprising an accelerator can be represented by concatenating the maps corresponding to individual magnets
\begin{equation}
    \mathcal{M} = \mathcal{M}_n \circ \mathcal{M}_{n-1} \circ \ldots \circ  \mathcal{M}_{1}.
\end{equation}

Various methods for efficiently concatenating such maps and extracting physics parameters such as betatron tunes or amplitude detuning have been developed \cite{SLAC-PUB-9574}. An efficient way of constructing a neural network which could model dynamics in such a system is to represent each map in this chain by a set of polynomial neurons (layer) computing exactly the same Taylor map, and then link these layers into a deep network. The NN has as its input the phase space coordinates and propagating the inputs through each layer is fully equivalent to applying the corresponding map.

\begin{figure}
\includegraphics[width=0.4\textwidth]{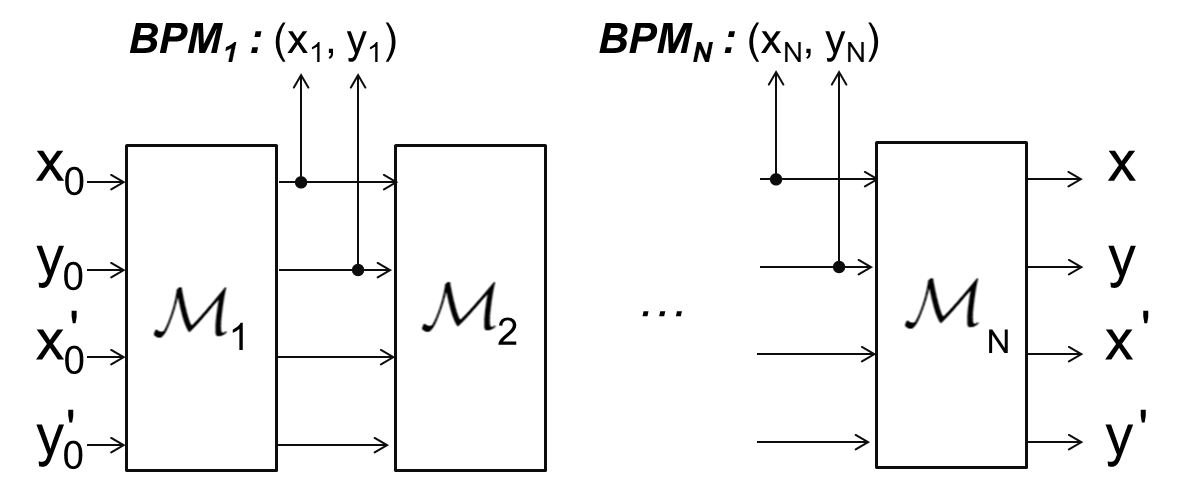}
\caption{Multi-output deep TM-PNN for particle accelerator.}
\label{fig.01}
\end{figure}

The general architecture of the TM-PNN is shown in Fig.~\ref{fig.01}. The network has a single input and can have an arbitrary number of outputs corresponding to observables. So, to have a beam trajectory at the beam position monitor (BPM) locations as output, the drift space containing the monitor is split into two parts at the monitor location. Two layers for drifts upstream and downstream of the monitor are created, and the upstream layer additionally has beam position coordinates $x$ and $y$ as outputs. Generally, the network output can be an arbitrary function of the outputs of all neurons in the network. Moreover, any number of parameters of the accelerator lattice such as the strength of quadrupole magnets can be trivially introduced. To add one additional parameter, the network input vector dimension is extended by one and the parameter becomes an additional input to the neurons where it enters.

For example, the transfer matrix for horizontal motion through a focusing quadrupole of length $L$
\begin{equation*}
M = \begin{pmatrix}
\cos(\sqrt{k}L)&\frac{1}{\sqrt{k}}\sin(\sqrt{k}L)\\
-\sqrt{k}\sin(\sqrt{k}L)&\cos(\sqrt{k}L)\\
\end{pmatrix}\\
\end{equation*}
can be parametrized by Taylor map and 
can be represented as a polynomial neuron with three-dimensional input $\XX_0 = (x_0, x'_0, k)$ and two-dimensional output $\XX = (x,x')$ where the weights up to the second order are
\begin{equation}
\begin{split}
W_1 &= \begin{pmatrix}
         1&L&0\\
         0&1&0\\
     \end{pmatrix},\\
W_2 &= \begin{pmatrix}
        0&0&-0.5L^2&0&0&0\\
        0&0&-L&0&-0.5L^2&0\\
    \end{pmatrix}.
\end{split}
\end{equation}

\begin{figure}
\includegraphics[width=0.48\textwidth]{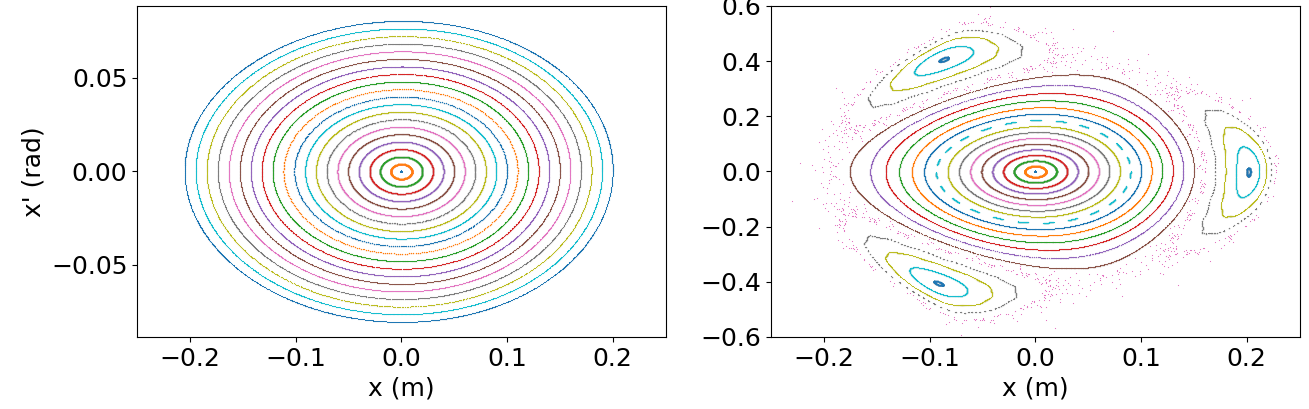}
\caption{Initialized from maps the TM-PNN accurately represents the parametric dependency of dynamics on magnet strength, such as 
the appearance of a third-integer resonance.}
\label{fig.03}
\end{figure}

We further illustrate the concept by an example of a network that corresponds to a simple but practically relevant magnet arrangement, the FODO structure with sextupoles, which exhibits nonlinear behavior typical of circular accelerators. The magnetic lattice and the corresponding NN are shown in   Fig.~\ref{fig.02}. This Taylor map-based polynomial neural network (TM-PNN) has 12 layers, for each component in the lattice. The resulting NN fully represents the dynamics of the lattice, including parametric dependency on quadrupole gradients. The system is nonlinear and exhibits resonant behavior. Fig.~\ref{fig.03} shows the phase space portraits of this lattice computed by TM-PNN. 

\section{NN training and symplectic regularization}

A TM-PNN pre-initialized with the Taylor map coefficients extracted from the ideal magnetic lattice already provides an accurate model of the ideal accelerator and does not require training. Training may be required if it is desired to fine-tune the model based on experimental data. The training is done in the usual way by fine-tuning the network weights given a set of inputs and outputs. For training, we use the Adam \cite{kingma2014method} algorithm with gradient clipping. The examples of such training are given in the last section of this paper.

The initial network weights are derived from the maps and are guaranteed to be physically meaningful. During the network training process, we need to ensure the physical consistency of the weights, so that they are changed in such a way that the layers still correspond to Taylor maps of magnets. The Taylor maps for the systems of ordinary differential equations contain weights that vary by orders of magnitude, and the often-used L1- or L2-norm-based regularizations of the weights during the training which try to reduce the absolute magnitude of the weights cannot be applied.

It is however possible to introduce physical invariants of the underlying dynamical systems as regularization penalties on the weights of the TM-PNN. For Hamiltonian systems representing single-particle beam dynamics, this can be the symplecticity \cite{arnold1989mathematical}, which means that for a map $\mathcal M: \XX_0 \rightarrow \XX$ 
\begin{equation}
\label{eq:sympl}
    \left(\frac{\partial \XX}{\partial \XX_0}\right)^T\,J\;\frac{\partial \XX}{\partial \XX_0} - J = 0,\;\forall \XX_0,\;J = \begin{pmatrix}0&I\\-I&0\end{pmatrix},
\end{equation}
where $I$ is an identity matrix, and $T$ is the matrix transpose. The symplectic property \eqref{eq:sympl} for TM-PNN leads to algebraic constraints on weights $W_i = \{w_i^{jk}\}$. The symplectic penalty can be implemented for an arbitrary Taylor map as an additional term for the loss function:
\begin{equation*}
    Loss = ME(\XX_{true}, \XX_{pred}) + S(W_1, \ldots, W_k),
\end{equation*}
where $ME$ is a metric that controls accuracy on training data, and $S$ is a symplectic reglarization penalty. For example, for the second-order Taylor map for two-dimensional state vector with $W_1=\{w_1^{ij}\},\;W_2=\{w_2^{mn}\}$ where $i,m$ are row and $j,n$ column indices in condition \eqref{eq:sympl}, the symplectic constraint becomes
\begin{equation}
\label{eq:algsympl}
    \begin{split}
    &w_1^{11}w_1^{22} - w_1^{12}w_1^{21} - 1 = 0,\\
    &w_2^{11}w_2^{23}-w_2^{13}w_2^{21} = 0,\\
    &w_2^{12}w_2^{23} - w_2^{13}w_2^{22} = 0,\\
    &w_1^{11}w_2^{22}-w_1^{21}w_2^{12} + 2w_1^{22}w_2^{11} - 2w_1^{12}w_2^{21} = 0,\\
    &w_1^{22}w_2^{12}-w_1^{12}w_2^{22} + 2w_1^{11}w_2^{23} - 2w_1^{21}w_2^{13} = 0.\\
    \end{split}
\end{equation}

The penalty $S$ is the sum of squares of all left-hand-side terms in \eqref{eq:algsympl}. Since this penalty does not depend on the inputs, the Hamiltonian structure is preserved for all new inputs which has a large impact on generalization. If the symplectic regularization is not considered during the training, the tuning of the maps leads to overfitting of the model that causes inaccurate predictions.

\begin{figure}[ht]
\includegraphics[width=0.38\textwidth]{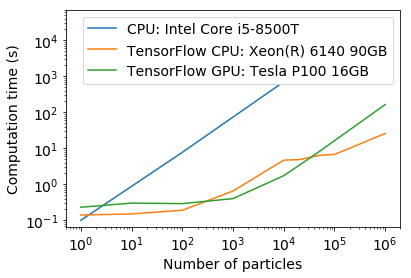}
\caption{Scaling of the computation time of the TM-PNN for one-turn tracking  in PETRAIII lattice.}
\label{fig.04}
\end{figure}

\section{Application to beam dynamics simulations}

TM-PNNs can be used for fast simulations of beam dynamics \cite{LieTF}. We implemented translation of beam optics models to TM-PNN using maps of up to second order for individual elements computed in OCELOT \cite{Agapov:2014yku}. From that point on, the computations necessary for particle tracking are completely performed by standard and highly optimized NN software such as TensorFlow \cite{TF}. This makes it possible to have highly efficient particle tracking on various parallel architecture without any additional effort. The computations were cross-checked on the PETRA III and PETRA IV lattices. Fig.~\ref{fig.04} presents the scaling of computational speed for four-dimensional particle tracking for the PETRA III lattice. On multi-core architectures, tracking of thousands of particles can be performed at the same speed as tracking of a single particle. Note that the network can deal with large lattices. So, the PETRA III lattice has 1519 elements, each of which was represented by a NN layer.

\section{Data-driven tuning of the TM-PNN}

Due to alignment errors, field imperfections, and other factors, ideal beam optics models of accelerators differ to various extents from the parameters of real machines. Adjustment of optics parameters in operation is normally performed. Usual methods of optics measurement and correction \cite{Tomas:2017zzu} exploit linear or weakly nonlinear dependency of observable parameters such as beam orbits on free model parameters such as magnet strengths and alignments. The problem of optics measurement is formulated as a least-squares minimization of free parameters given a set of observations, and then a correction is done using some of these parameters as controls to minimize the discrepancy between theoretical and observed models. 

One of the general shortcomings of these methods is that the number of free parameters such as offsets and strength errors of magnets is larger than the number of observables, which is ultimately limited by the number of free parameters in the phase space transformation between adjacent beam position monitors. For example, for linear coupled motion the number of such parameters is the dimension of the group of 4D symplectic matrices which is  $n(2n+1)=10$ so that the number of possible observable parameters in the beam optics is $10 N_{BPM}$. So, for the PETRA III ring at DESY the number of magnets is $N_{mag} = 1519$ and $N_{BPM}=246$. With 6 free parameters for geometrical misplacement of the components, optics determination becomes strictly speaking ill-posed. Employed least-square fit usually has a different number of parameters than intrinsically present in the model, and additional assumptions or empirical knowledge is often required to successfully use such fits. In contrast, a TM-PNN with a minimal set of layers (i.e. one layer for the transfer map between two BPMs) does not have the shortcomings mentioned above.

A TM-PNN can serve as the backbone for both optics measurement and for beam control. The number of layers should be the smallest possible but include all observables. For example, maps for all magnets between BPMs can be concatenated. However, elements used for beam control should be preserved as separate layers.

The first step, model inference, is fitting all the weights in the network for a given training set of observations, such as orbit measurements. The second step, control, consists of giving the network a target function, such as the target orbit, as a new training dataset, and then using only the weights in the layers corresponding to control elements to fit the desired target. Since by construction of the network there is a unique correspondence between the weights and control parameters, the appropriate correction can be directly inferred. The TensorFlow engine can be used for both training of the network with available observations, and for finding the control parameters. This concept is demonstrated below on several examples.

\begin{figure}
\includegraphics[width=0.5\textwidth]{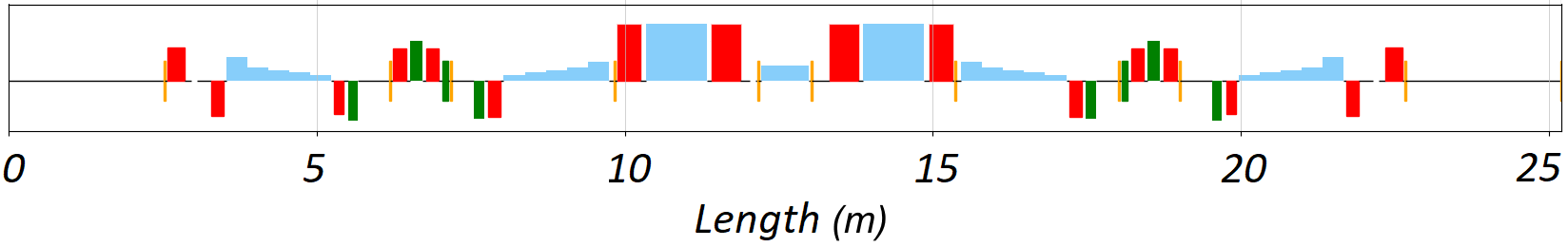}
\caption{The PETRA IV cell (a seven-bend achromat of ESRF-EBS type) that consists of bending magnets, shown in blue, focusing  and defocusing quadrupoles, shown in red, and focusing (sf) and defocusing (sd) sextupoles, shown in green.}
\label{fig.05}
\end{figure}

\subsection{Simulated data for PETRAIV}
First, we used a simplified version of the Multi-Bend Achromat lattice of PETRA IV (see Fig.~\ref{fig.05}) \cite{Schroer:ig5056} to model single-pass trajectory correction with a  TM-PNN. 
Single-pass trajectory correction is an important step in machine startup when beam accumulation is not yet achieved.
Random magnet misalignments were generated, and single-pass beam trajectories simulated. In the following we give an example of trajectory correction on the example of a single cell.
The cell has 36 magnets and a total of 166 logical elements (free spaces, markers, etc.). For simplicity, we consider only beam motion in horizontal plane $x-x'$. The cell has 11 beam position monitors (BPMs) and 10 orbit correctors.

\begin{figure}
\includegraphics[width=0.45\textwidth]{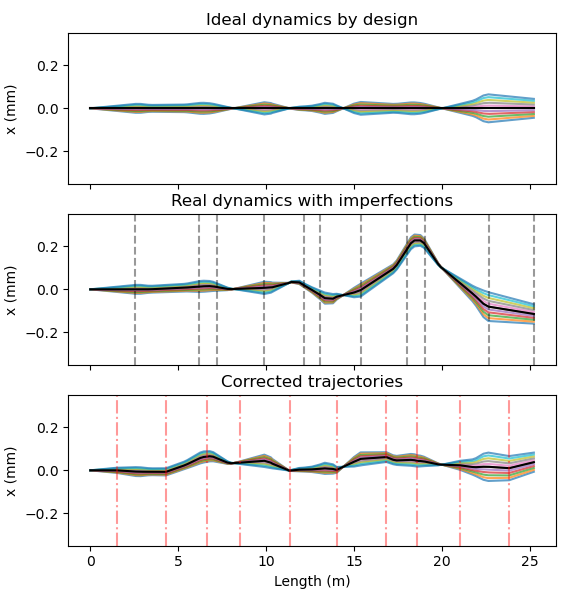}
\caption{Trajectory correction in a Multi-Bend achromat lattice: Trajectories in an ideal lattice (top), in a lattice with random misalignments (middle), and after correction by a TM-PNN (bottom). BPM and corrector positions are shown with black and red dashed lines respectively.}
\label{fig.06}
\end{figure}

The orbit correction is formulated as a nonlinear optimization problem
\begin{equation}
\label{eq:orcbit_cor}
    F = \sum_{i=0}^{10}|| x_i (c_0, c_1, \ldots, c_9) || \rightarrow 0,
\end{equation}
where $x_i$ is a beam coordinate measured at $i$-th monitor, $c_j$ is a strength of $j$-th corrector magnet. The simulated trajectory including misalignments is fed into the TM-PNN representing the ideal lattice to re-calibrate the model.The network was then used to find corrector strength resulting in zero trajectory. The correction results for one achromat cell are shown in Fig.~\ref{fig.06}. Only one trajectory is required to train the network which can then perform the orbit correction.

\subsection{Experiments at PETRAIII}
Experiments were performed at PETRA III to demonstrate applicability of TM-PNN for single-pass trajectory correction in real operation. 
As mentioned before, single-pass trajectory correction is an important task in commissioning and startup of accelerators where the optics model can also be subject to uncertainties. Moreover, diagnostics resolution in single-pass mode is worse than for the stored beam and the demonstration of model re-calibration thus becomes more challenging. 
For the measurement the beam position monitors were set up to read single-turn data and were triggered at beam injection into the ring.
The first orbit corrector in the storage ring upstream from the injection point was set up such that the beam was making only one turn.
The TM-PNN model was initialized with the nominal PETRA III lattice, and the one-turn trajectories consisting of $(x,y)$ positions at 246 BPM locations were used for training.

\subsubsection{Beam threading}
\begin{figure}
\includegraphics[width=0.48\textwidth]{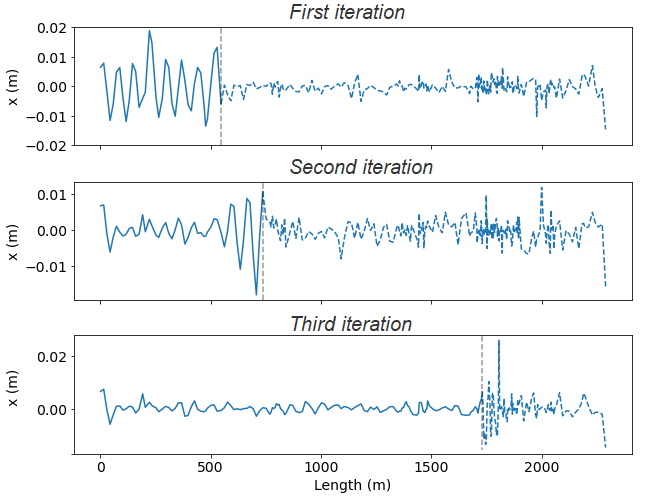}
\caption{Several iterations of beam threading with a TM-PNN. }
\label{fig.07}
\end{figure}

The first experiment demonstrated that a TM-PNN can be used for beam threading through the machine and for orbit correction. 
We switched off all corrector magnets which leads to the beam being able to travel through only a part of the ring. To propagate the beam further we perform orbit correction with the available (upstream from the beam loss location) correctors by the TM-PNN. 
For example, only 30 correctors are available before beam is lost in the first iteration, with the total amount of 210 horizontal and  194 vertical correctors in the ring. Given the measured data at each step, the TM-PNN predicts values of corrector magnets that will reduce displacement of the measured orbit. These corrector values are applied to the machine and a new measurement is acquired. After several steps, the beam is propagated along the ring. Fig.~\ref{fig.07} shows several iterations of beam threading and trajectory correction with a TM-PNN. Solid lines represent beam coordinates, dashed lines are noise downstream from the beam loss. 
Note that in this example model re-calibration is not essential due to relatively weak nonlinearities in the machine and the primary goal was the demonstration of orbit control with the help of the TM-PNN.

\subsubsection{Optics imperfection}
\begin{figure}[ht]
\includegraphics[width=0.35\textwidth]{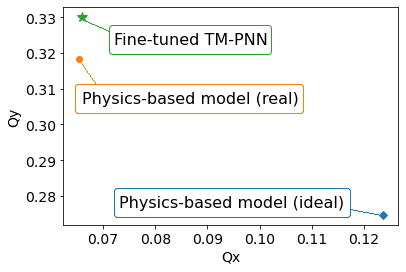}
\caption{Tunes in the horizontal(Qx) and vertical (Qy) planes.}
\label{fig.08}
\end{figure}

In the second experiment we used the same beam set-up to demonstrated model re-calibration on partial, noisy, and limited observations.
Optics imperfections were introduced in a controlled way by detuning the strength of one quadrupole magnet by 20\%. The network was trained on the single-pass trajectories from this set-up and then used in multi-turn tracking.
Since the imperfection is known, we independently reproduced that optics in a traditional tracking code (OCELOT).
Multi-turn particle tracking was done in both codes and the tunes extracted by Fourier analysis (see Fig.~\ref{fig.08}).
Note that only a single one-turn trajectory was used for model re-calibration. Tune measurement based on such data with traditional methods is generally impossible.

\section{Supplementary material}
The implementation of the TM-PNN architecture in Keras/TensorFlow and the described examples of a FODO lattice, simulations of the PETRA IV cell, and translating of the PETRAIII lattice to the TM-PNN architecture are available under \url{https://github.com/andiva/TM-PNN}

\;

\begin{tabular}{ll}
\textbf{/tm\_pnn:} & implementation of the TM-PNN in \\
& Keras with TensorFlow backend\\
& with several symplectic loss functions\\
\textbf{/demos/}&\\
\textbf{\;\;\;\;FODO}:  & FODO lattice with resonances\\
\textbf{\;\;\;\;PETRAIV}:  & orbit correction for simulated data\\
\textbf{\;\;\;\;PETRAIII}:  & conversion  of the PETRA III with 1519\\
&magnets into TM-PNN\\
\end{tabular}

\section{Conclusion}

We introduced a new polynomial neural network architecture that is ideally suitable for representing beam dynamics in charged particle accelerators. The networks are already pre-initialized to represent linear and nonlinear beam dynamics and can be effectively trained using the symplectic regularization technique we proposed. They provide both a backbone for fast parallel computations and a universal ML-based approach for experimental measurement and correction of charged particle beam optics. 
Symplectic regularization is limited to hamiltonian systems, but the TM-PNN approach is a more general. In \cite{TMPNN} it is shown that an arbitrary system of nonlinear ODEs can be translated to Taylor maps. So, radiation and/or space charge effects can be introduced into TM-PNN by, for example, mapping onto them the systems of equations from \cite{ASC}. 
Also, further research on the estimation of accuracy, performance, and limitations of the proposed method as well as work on introducing it to the daily accelerator operation should be performed.

\section{Acknowledgements}

The authors are grateful to Gero Kube, Joachim Keil, Gajendra Kumar Sahoo  and Rainer Wanzenberg for support and active participation in measurements at the PETRAIII storage ring. The work was supported by the AMALEA Helmholtz Innovation Pool Project.

\bibliography{paper}

\end{document}